\documentclass[letterpaper, 10 pt, conference]{ieeeconf}  

\IEEEoverridecommandlockouts                              

\usepackage{color}
\usepackage{graphicx}
\usepackage[table,xcdraw]{xcolor}
\usepackage{pifont}
\usepackage{cite}
\usepackage{url}
\usepackage[colorlinks=true,linkcolor=red,citecolor=green]{hyperref}
\usepackage[caption=false]{subfig}

\title{\LARGE \bf
Disambiguate Gripper State in Grasp-Based Tasks:\\Pseudo-Tactile as Feedback Enables Pure Simulation Learning
}

\author{Yifei Yang$^{1}$, Lu Chen$^{1}$, Zherui Song$^{1}$, Yenan Chen$^{1}$, Wentao Sun$^{2}$, \\ Zhongxiang Zhou$^{1}$, Rong Xiong$^{1}$ and Yue Wang$^{1}$
\thanks{$^{1}$Yifei Yang, Lu Chen, Zherui Song, Zhongxiang Zhou, Rong Xiong and Yue Wang are with the State Key Laboratory of Industrial Control and Technology, Zhejiang University, Hangzhou 310027, China. Yue Wang is the  corresponding author
        {\tt\small wangyue@iipc.zju.edu.cn}}%
\thanks{$^{2}$ Wentao Sun is with Huawei.}%
}

\begin{document}

\maketitle
\thispagestyle{empty}
\pagestyle{empty}

\begin{abstract}

Grasp-based manipulation tasks are fundamental to robots interacting with their environments, yet gripper state ambiguity significantly reduces the robustness of imitation learning policies for these tasks. Data-driven solutions face the challenge of high real-world data costs, while simulation data, despite its low costs, is limited by the sim-to-real gap. We identify the root cause of gripper state ambiguity as the lack of tactile feedback. To address this, we propose a novel approach employing pseudo-tactile as feedback, inspired by the idea of using a force-controlled gripper as a tactile sensor. This method enhances policy robustness without additional data collection and hardware involvement, while providing a noise-free binary gripper state observation for the policy and thus facilitating pure simulation learning to unleash the power of simulation.  Experimental results across three real-world grasp-based tasks demonstrate the necessity, effectiveness, and efficiency of our approach. Videos are available on \href{https://yifei-y.github.io/project-pages/Pseudo-Tactile-Feedback/}{Project Page}.

\end{abstract}

\section{Introduction}

Grasp-based manipulation spans a wide range of tasks, from simple pick-and-place \cite{pnp1, pnp2} to more complex interactions like tool usage \cite{tool-usage-1, tool-usage-2}, making it a fundamental capability for robots to engage with the environment. One promising way to teach robots these skills is imitation learning (IL) \cite{act, dp}, which enables robots to learn directly from expert demonstrations through supervised learning.

\begin{figure}[thpb]
\centering
\subfloat[Gripper state ambiguity makes policies vulnerable to disturbances.]{\includegraphics[width=0.48\textwidth]{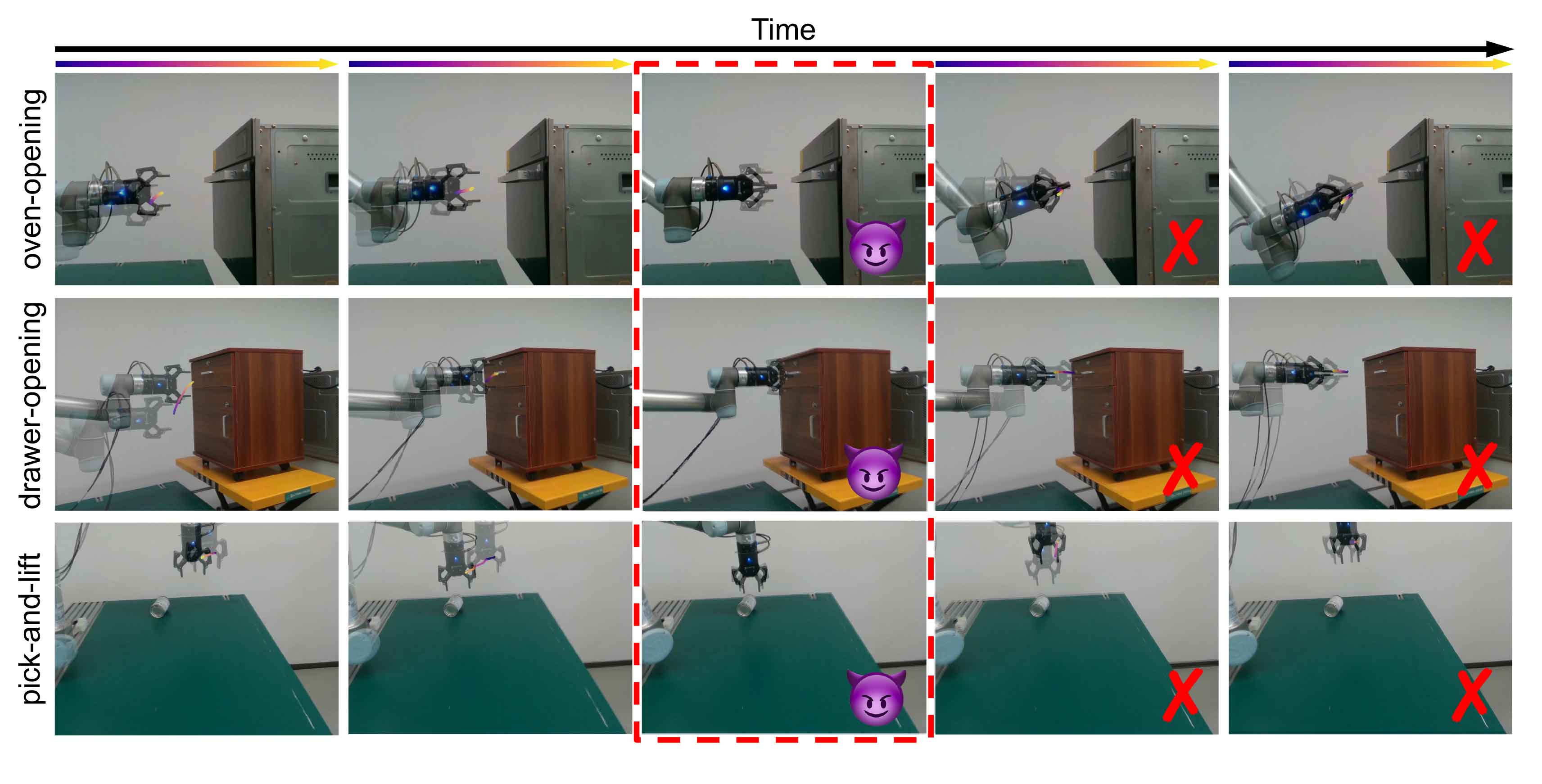}%
\label{fig: teaser-1}}
\hfill
\subfloat[We propose pseudo-tactile feedback to disambiguate gripper state.]{\includegraphics[width=0.45\textwidth]{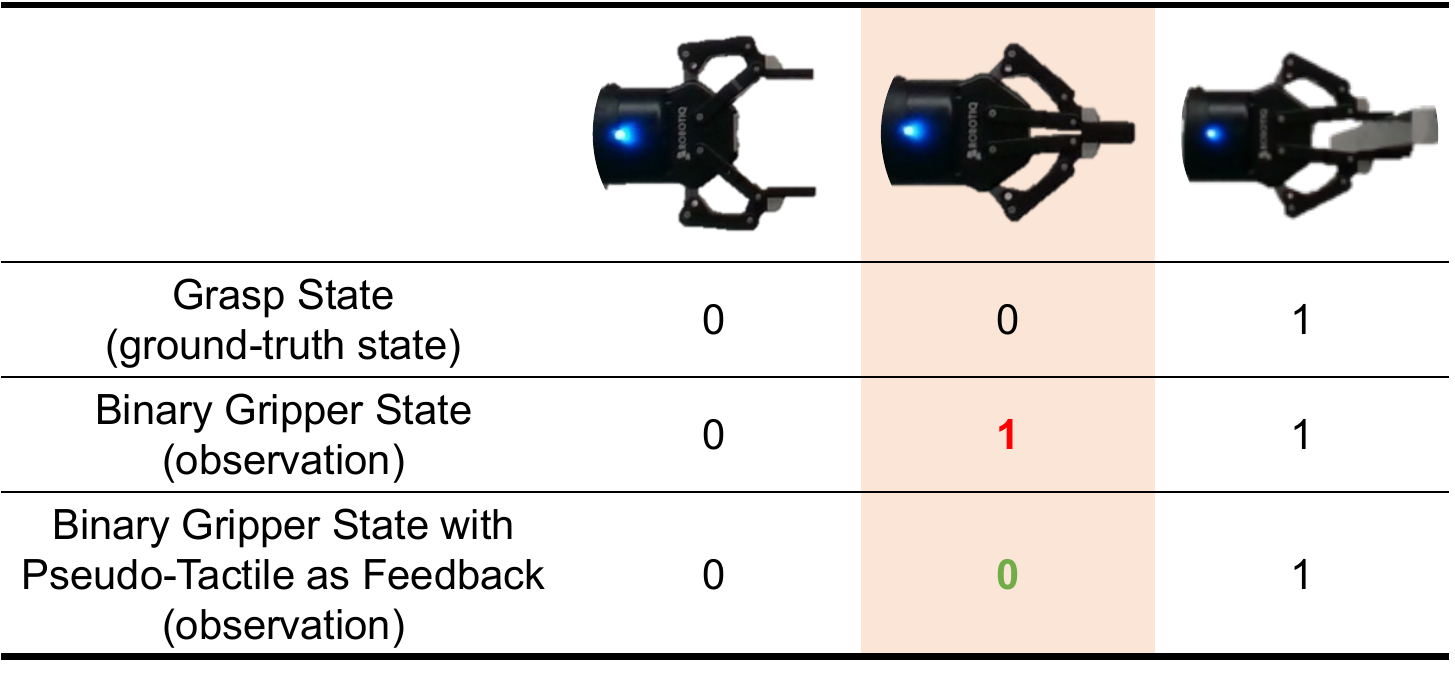}%
\label{fig: teaser-2}}
\caption{(a) Policies for grasp-based tasks are susceptible to gripper disturbance (forcing the gripper to close at the moment marked by the red dashed box). While the robot does not grasp the object, the policy outputs the action command for the post-grasp phase. (b) We propose pseudo-tactile feedback to accurately reflect the grasp state, effectively disambiguating the gripper state.}
\label{fig: teaser}
\vspace{-0.6cm}
\end{figure}

The efficacy of IL is heavily dependent on the quantity and quality of the provided demonstrations. However, it is challenging to account for unexpected disturbances during data collection, such as an object slipping from the gripper during transport or a handle slipping from the grasp when opening an oven. We observed that these incidents can be generalized as cases where the gripper closes without successfully grasping the target object. We simulate these scenarios by forcing the gripper to close before reaching the grasp pose in our experiments. The results reveal that IL policies are not robust to such disturbances. Take the task of opening an oven as an illustrative example. When subjected to this perturbation, the policy failed to recognize the unsuccessful grasp and instead transitioned directly to the post-grasp stage, outputting a “pull” action despite not having secured the handle. As shown in Fig. \ref{fig: teaser-1}, similar failures occur in the drawer-opening and pick-and-lift tasks, where the policy prematurely moves to the post-grasp stage rather than reattempting the grasp when disturbed.

We attribute this to gripper state ambiguity and data bias. As shown in Fig. \ref{fig: teaser-2}, the binary gripper state cannot distinguish between a successful and empty grasp. In demonstrations, a closed gripper always indicates a successful grasp, but this correlation breaks under perturbations during inference. While allowing the policy to observe continuous gripper joint angles and collecting more real-world data to cover diverse grasp outcomes could help, actively inducing grasp failures during teleoperation is impractical, as it requires extensive human intervention and disrupts natural data collection. Moreover, the expanded observation space and additional randomization would make data collection costs surge. Simulation offers an efficient alternative for data collection, but the sim-to-real gap hinders its potential. The gripper joint angle in simulation when grasping an object differs from it in the real world due to the imperfections in simulation assets and the physics engine, such as geometric artifacts and the interpenetration issue, making it difficult to use the gripper joint angle as an observation in simulation.

Let’s delve deeper into this issue. Why is it that data collected through teleoperation like UMI \cite{umi} typically associates a closed gripper state with a successful grasp? This is because human feedback, such as tactile feedback, is involved during data collection and helps to ensure the correspondence. However, this feedback is unavailable to the policy, leaving only the ambiguous feedforward binary gripper state, i.e., the current gripper state observation depends on the previous gripper control command rather than reflecting the grasp state. Therefore, an intuitive way to disambiguate gripper state would be to integrate a tactile sensor to provide feedback. However, incorporating tactile sensors presents several challenges, such as the complexity of integration, the need for calibration, real-time processing requirements, and increased hardware costs.

The principle of tactile sensors is that they convert the deformation of sensor material - when in contact with an object and achieving force equilibrium - into measurable changes in resistance, capacitance, or visual signals, which reflect tactile information. Our key insight is that a force-controlled two-finger gripper can be viewed as a tactile sensor. The two fingers act as the deformable material of the tactile sensor. When the gripper grasps an object and reaches force equilibrium, the deformation of the fingers, represented by the gripper joint angle, provides \textit{pseudo-tactile} information.

In this paper, we employ pseudo-tactile as feedback to disambiguate the gripper state in grasp-based tasks without additional data collection and hardware involvement. With pseudo-tactile information from the force-controlled gripper, we design a simple yet effective closed-loop gripper controller that ensures the closed gripper state always corresponds to a successful grasp, as illustrated in Fig. \ref{fig: teaser-2}. After disambiguating the gripper state, there is no need for the policy to use the continuous gripper joint angle as an observation. Instead, it can use the noise-free binary gripper state, enabling pure simulation learning. Therefore, we can benefit from the advantages of simulation, including access to the privileged state of the environment, painless domain randomization, and so on. 
We conducted experiments on three real-world grasp-based tasks. The results demonstrate that, with pseudo-tactile as feedback, the policy can effectively handle the aforementioned disturbances. Furthermore, the policy trained on pure simulation data outperforms baseline models utilizing real-world teleoperation data. Our key contributions are summarized as follows:

\begin{itemize}
    \item We propose a novel pseudo-tactile as feedback approach to disambiguate gripper state in grasp-based tasks, enhancing the system's robustness without additional data collection and hardware involvement.
    \item The proposed approach allows the use of a binary gripper state, which enables policy learning with pure simulation data, and thus we can take advantage of the various benefits of simulation.
    \item We conduct experiments across three real-world grasp-based tasks, and the results highlight the effectiveness of the proposed approach.
\end{itemize}

\section{Related Works}

\subsection{Imitation Learning for Grasp-Based Manipulation}

In recent years, researchers have made significant progress in imitation learning. Some works have evolved policy structures. For example, Zhao et al. \cite{act} proposed ACT, employing a transformer architecture to predict a sequence of action tokens from encoded image tokens. Diffusion Policy \cite{dp} uses the diffusion process to model a multimodal conditional action distribution. Some other works focused on fusing multimodal information, such as 3D perception \cite{dp3}, language instruction \cite{rt1, rt2}, and audio \cite{audio}, to improve the representation capability.

Many researchers have adopted the imitation learning paradigm to enable robots to learn grasp-based manipulation tasks, not only in the most direct tasks like pick-and-place \cite{il-pnp-1, il-assembly-2} but also in more complex tasks such as assembly \cite{il-assembly-1, il-assembly-2} and articulated object manipulation \cite{il-articulated-2, unidoormanip}. One of the most notable advances is the development of robotic foundation models, particularly Vision-Language-Action Models (VLA) \cite{rt1, rt2, gr-1, gr-2, rdt, pi_0}, which achieved remarkable generalization and robustness by leveraging massive datasets, enabling robots to perform a wide variety of grasp-based manipulation tasks. As a result, significant efforts have been dedicated to creating large-scale robotic manipulation datasets, such as Open X-Embodiment (OXE) \cite{oxe} and DROID \cite{droid}.

However, enhancing policy robustness with data-driven approaches is costly in the real world, and the sim-to-real gap hinders the potential of simulation. We analyzed the reasons why imitation learning policies for grasp-based tasks are susceptible to gripper disturbance and proposed a novel method to enhance the policy’s robustness without the need for additional data collection.

\subsection{Manipulation with Tactile}

Tactile information plays a crucial role in human-environment interactions. Inspired by this, many researchers have employed tactile sensors to enhance robotic manipulation. Lin et al. \cite{tactile-1} developed a teleoperation system that can record tactile information and proposed a visuotactile policy for bimanual manipulation. Huang et al. \cite{tactile-2} fused visual and tactile information into a unified 3D representation space, leading to a more effective policy learning process. Despite their effectiveness, incorporating tactile sensors comes with trade-offs, such as increased costs and system complexity. Instead, we use a force-controlled gripper to provide pseudo-tactile signals, benefiting from tactile information without introducing additional hardware.

\section{Gripper State Ambiguity} \label{sec: 3}

We start by explaining what gripper state ambiguity is, why it occurs, and what it leads to. 

\textbf{What is gripper state ambiguity?} 
In robotic systems, the state refers to the complete and accurate description of the system’s internal and external variables, encompassing all necessary information for decision-making, while the observation represents the partial or noisy measurement of the state obtained through sensors, which may not fully reflect the true system state. Whether the robot has successfully grasped an object is a crucial state for a control system designed to perform grasp-based tasks, denoted as \textit{grasp state} in \ref{fig: teaser-2}. This state determines whether the policy should actuate the robot to continue attempting to grasp or transition to the post-grasp phase. In demonstrations, the observation of a closed gripper typically corresponds to a successful grasp, causing the policy to overfit this relationship. As such, the \textit{gripper state} serves as the primary observation used to measure the \textit{grasp state} (it is important to note that the \textit{gripper state} is an observation, whereas the \textit{grasp state} is a state). However, during inference, external disturbances or errors may cause the robot to close the gripper without successfully grasping the object. While the grasp state differs, the primary observation remains unchanged, as illustrated in Fig. \ref{fig: teaser-2}. Therefore, the observation of the gripper state is not a reliable measurement for the grasp state, which is what we refer to as gripper state ambiguity.

\textbf{Why does gripper state ambiguity occur?}
The root cause of gripper state ambiguity lies in the fact that the policy learns the mapping from observation to action based on human-teleoperated demonstrations, but the observation information available to the policy differs from what the human operator perceives. During data collection, humans do not rely on the gripper state to determine whether the object has been successfully grasped. Instead, they use tactile feedback, which serves as a reliable observation of the grasp state. Unfortunately, the policy does not have access to tactile feedback and mistakenly treats the feedforward gripper state as a reliable indicator of the grasp state. This mismatch leads to the occurrence of gripper state ambiguity.

\begin{figure}[thpb]
\centering
\includegraphics[width=0.47\textwidth]{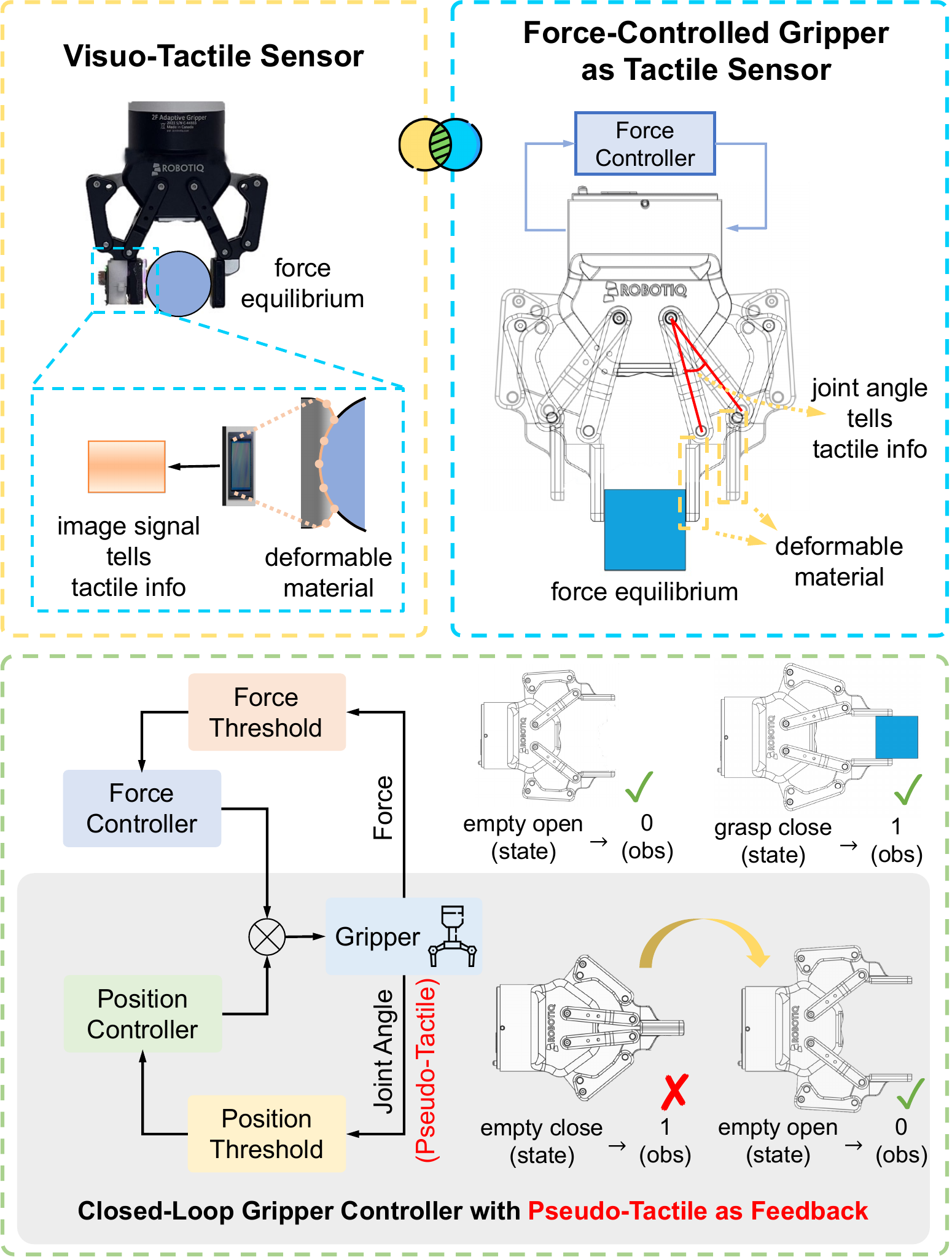}
\vspace{-0.3cm}
\caption{A force-controlled gripper can act as a tactile sensor providing pseudo-tactile information. We designed a closed-loop gripper controller with pseudo-tactile information as feedback to convert the \textit{empty close} state into the \textit{empty open} state, disambiguating gripper state.}
\label{fig: pseudo_tactile}
\vspace{-0.6cm}
\end{figure}

\textbf{What will gripper state ambiguity lead to?}
The correspondence between a closed gripper and a successful grasp is easily broken by disturbance, making the policy vulnerable to disturbance. As illustrated in Fig. \ref{fig: teaser-1}, the policy outputs a “pull” or "lift" action as soon as the gripper closes, regardless of whether the object is grasped.

\section{Methodology}

Building on the insight that a force-controlled gripper can act as a tactile sensor, we propose a pseudo-tactile feedback approach that can disambiguate the gripper state in grasp-based tasks, enhancing the robustness of the imitation learning policy without additional data collection and hardware involvement. Having disambiguated the gripper state, policies can use the noise-free binary gripper state as an observation, which enables pure simulation learning. Therefore, we can benefit from the advantages of simulation. In Sec. \ref{sec: 4.1}, we present the implementation of pseudo-tactile feedback. In Sec. \ref{sec: 4.2}, we detail the process of training the policy using pure simulation data and transferring it to the real world.

\subsection{Pseudo-Tactile Feedback} \label{sec: 4.1}

As illustrated in Sec. \ref{sec: 3}, the essential reason for gripper state ambiguity is the unavailability of tactile feedback to the policy. While this issue could be addressed by introducing tactile sensors, doing so would increase both the cost and complexity of the system.

\textbf{Force-Controlled Gripper $\approx$ Tactile Sensor.} The basic principle of tactile sensors is that when the sensor is in contact with an object and reaches force equilibrium, the sensor converts the deformation of the sensor material into a measurable change in resistance, capacitance, or visual signals that reflect tactile information, as shown in Fig. \ref{fig: pseudo_tactile}. Based on this principle, a force-controlled gripper, such as Robotiq 2F-85, can be viewed as a tactile sensor, with its two fingers as the deformable material. When the gripper grasps an object and reaches force equilibrium, the deformation of the fingers can be converted into a measurable change in the gripper joint angle, providing pseudo-tactile information.

\textbf{Gripper Controller Achieves Pseudo-Tactile Feedback.} The grasp state can be categorized into three possible conditions: \textit{empty open} (the gripper is open and has not grasped an object), \textit{grasp close} (the gripper is closed and grasping an object), and \textit{empty close} (the gripper is closed but has failed to grasp an object). However, the binary gripper state observation is identical (i.e., 1) for both the \textit{grasp close} and \textit{empty close} states, leading to ambiguity. As illustrated in Fig. \ref{fig: pseudo_tactile}, we design a simple yet effective low-level closed-loop gripper controller with pseudo-tactile information as feedback, which can convert the \textit{empty close} state into the \textit{empty open} state, thereby eliminating the ambiguity. The controller takes a pseudo-tactile signal (i.e., the gripper joint angle when reaching force equilibrium) as input and outputs control commands for the gripper according to straightforward manually-designed rules. Specifically, when the gripper closes but the pseudo-tactile signal indicates that no object has been grasped (i.e., the gripper joint angle reaches its maximum value), the controller overrides the high-level policy’s command and forces the gripper to open, converting the \textit{empty close} state into the \textit{empty open} state. This ensures that the binary gripper state observation corresponds to 1 only when the gripper has successfully grasped an object (\textit{grasp close}), making it a reliable indicator of grasp success or failure.

Our approach offers several key advantages. First, it does not incorporate additional hardware, maintaining the cost-efficiency and simplicity of the system. Second, it avoids the expansion of observation space and the increase of randomization variables, which eliminates the demand for additional data and facilitates the policy learning.

\subsection{Pure Simulation Training} \label{sec: 4.2}

Collecting demonstrations via human teleoperation in the real world is time-consuming and labor-intensive. Simulation offers an efficient alternative by providing access to the privileged state of the environment, supporting painless domain randomization and environment resets, and leveraging parallel computation for acceleration. Nevertheless, a key limitation of simulation data is the sim-to-real gap. For example, imperfections in simulation assets and the physics engine, such as geometric artifacts and the interpenetration issue, cause discrepancies in the gripper joint angles when grasping the same object in the simulation and the real world. This makes it difficult to use the gripper joint angle as an observation in simulation. 

\begin{figure}[thpb]
\centering
\includegraphics[width=0.47\textwidth]{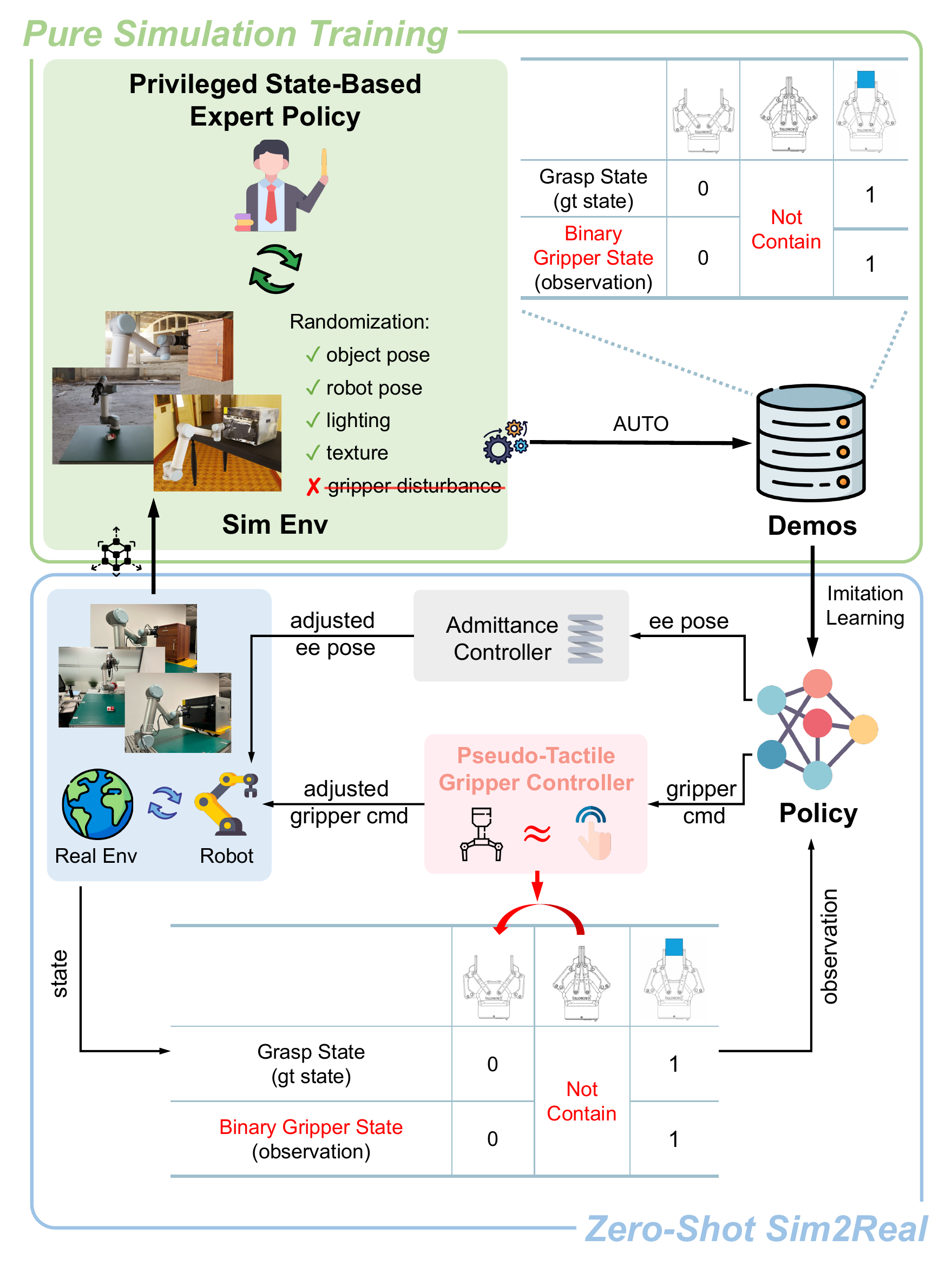}
\vspace{-0.3cm}
\caption{Once the gripper state is disambiguated, there is no need to perform gripper disturbance during data collection, and the policy can use the binary gripper state observation, enabling pure simulation training. We design a state-based expert policy for automatic data collection and apply real-to-sim techniques, randomization, and admittance control to mitigate the visual sim-to-real gap and reduce the stress caused by kinematic discrepancies.}
\label{fig: pure_sim}
\vspace{-0.6cm}
\end{figure}

With pseudo-tactile feedback converting the \textit{empty close} state into the \textit{empty open} state, real-world inference can align with simulation data as shown in Fig. \ref{fig: pure_sim}. The policy will not be bothered by the gripper state ambiguity, thus eliminating the need for gripper disturbance during data collection and the use of continuous gripper joint angle as an observation. Instead, the policy can use the noise-free binary gripper state as an observation, which diminishes the impact of the sim-to-real gap and allows us to train the policy using pure simulation data so that we can benefit from the advantages of simulation. Next, I will provide a detailed explanation of how we collect data in simulation and the techniques we use to mitigate the sim-to-real gap.

\textbf{Demo Generation.}
We abstract the three grasp-based tasks (pick-and-lift, drawer-opening, and oven-opening) into three stages: pre-grasp, grasp, and post-grasp. For the pre-grasp and grasp stages, we manually define one object-centric key pose $T_{\mathcal{OK}}$ respectively. Fetching the poses of objects of interest from the simulator $T_{\mathcal{WO}}$, we can compute the target poses in the world coordinate for the robot’s end effector $T_{\mathcal{WE}} = T_{\mathcal{WO}} \cdot T_{\mathcal{OK}}$. We then perform interpolation to generate trajectories from the current pose to the target poses. While more advanced planning algorithms could be employed, we find the simple interpolation is satisfactory in most cases, being able to cope with over 90\% randomization. For the post-grasp phase, we design motion rules for each task to satisfy kinematic constraints: 
\begin{itemize}
    \item Pick-and-Lift: make the robot's end-effector move vertically upward from the grasp position.
    \item Drawer-Opening: command the robot to pull back in a direction perpendicular to the cabinet.
    \item Oven-Opening: allow the robot to dynamically adjust the direction of pull-down according to the pose of the oven door, following a near-circular arc trajectory
\end{itemize}
The completion of rollouts is automatically determined by monitoring either the object poses or the joint angles, and we only keep successful ones as demonstrations.

\begin{figure*}[thpb]
\centering
\includegraphics[width=0.9\textwidth]{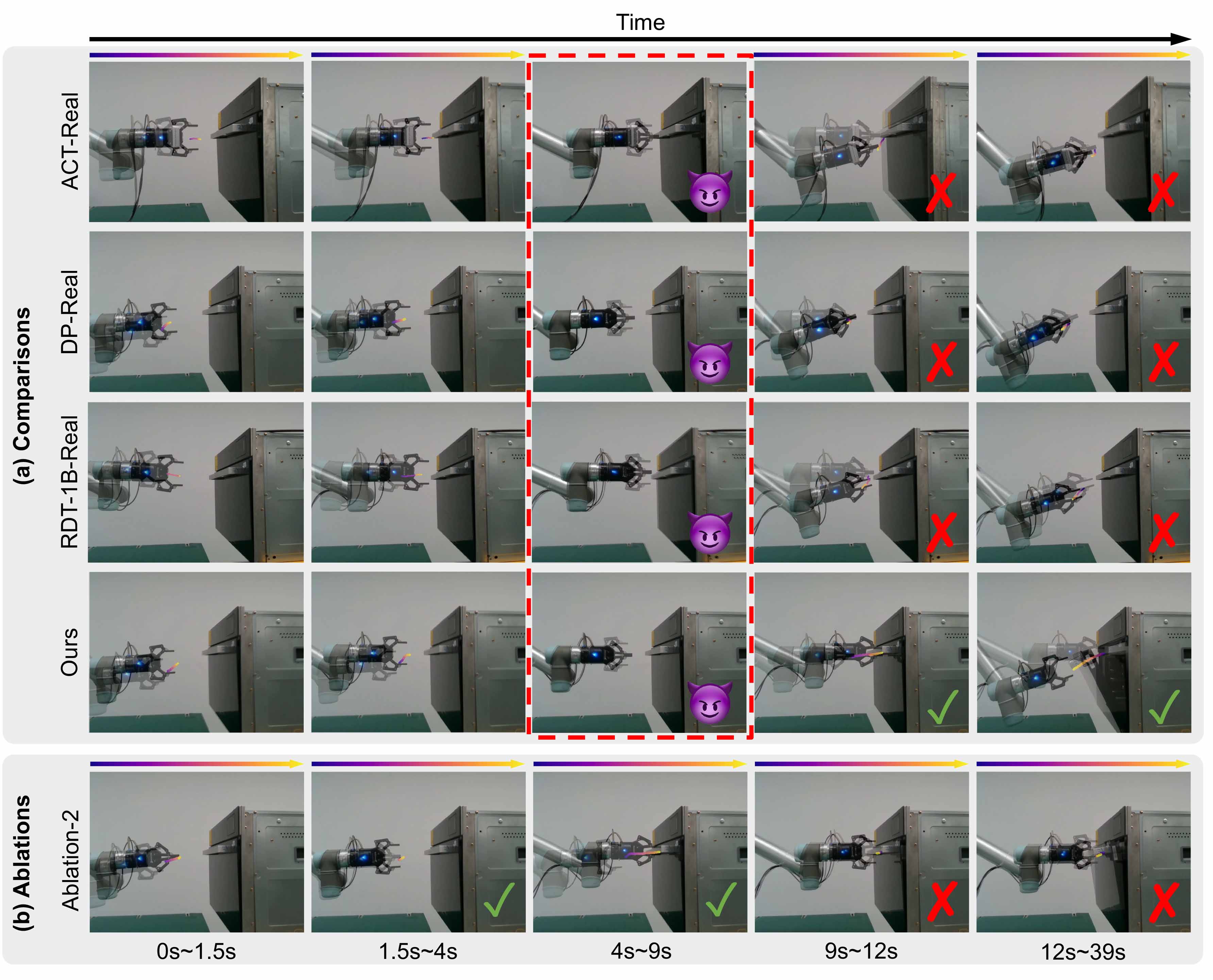}
\vspace{-0.4cm}
\caption{(a) Real-world oven-opening comparisons. We introduced a disturbance at the moment marked by the red dashed box by forcing the gripper to close. While the baseline models failed due to the disturbance, ours completed the task successfully. (b) One of rollouts of a policy trained using simulation data with gripper randomization in ablation studies. It successfully recovered from the empty grasp (1.5s-4s) and grasped the handle (4s-9s), but could not complete the task.}
\label{fig: qualitative_exp}
\vspace{-0.6cm}
\end{figure*}

This approach enables the collection of much larger and more diverse datasets in the same time than real-world teleoperation. For example, in 2 hours, we collected 1000 oven-opening demonstrations in simulation using a single GeForce RTX 4060 Ti, whereas real-world teleoperation yielded around 50 demonstrations over 3 hours. Using multiple GPUs for parallel computation can further accelerate the process. Each rollout applies randomizations (object and robot poses, lighting, and textures) to diversify the scenario. While setting up the simulation environment and expert policy requires human involvement, the data collection process is fully automated thereafter, meaning human effort does not scale with data volume. 

Moreover, simulation facilitates on-policy data collection, such as DAgger \cite{dagger}, where the policy is trained and simultaneously tested, with expert demonstrations used to correct failures and augment the dataset. This can improve the efficiency of data collection and policy learning. Such a method is difficult to implement in real-world scenarios, where manual intervention is required for each failed case.

\textbf{Sim-to-Real Transfer.}
The first step is to create realistic simulated assets. For objects with simple geometry and texture, like a wooden cabinet, we use 3D modeling software to create the geometric shapes and apply textures in Isaac Sim. For more complex objects, like an oven, we apply 3D reconstruction techniques to create textured meshes and then post-process them to add articulations. We also apply randomizations to diversify demonstrations, enhancing the policy’s robustness to the visual sim-to-real gap.

For tasks involving articulated objects with revolute joints, the kinematical sim-to-real gap often degrades policy performance. Discrepancies in joint axis positions can lead to excessive stress in the real world, triggering emergency stops and causing task failure. To address this, we employ admittance control \cite{admittance}, which allows the robot to adjust its motion in response to external force and torque, ensuring accurate tracking of the policy output trajectory $\left[\ddot{x}_d, \dot{x}_d, x_d\right]$ while providing compliance to external force and torque $F_{ext}$. Essentially, admittance control acts like adding a spring to the end effector: when no external force or torque is applied, the end effector follows the policy’s command; when external force and torque are present, the pose is adjusted by a correction offset based on the external force and torque. The dynamics model of admittance control is:
\vspace{-0.3cm}

\begin{equation}
    \label{equ:admittance}
    M\left(\ddot{x}_c-\ddot{x}_d\right)+D\left(\dot{x}_c-\dot{x}_d\right)+K\left(x_c-x_d\right)=F_{e x t}
\end{equation}
where $M$ is the mass matrix, $D$ is the damping matrix, $K$ is the stiff matrix, and $\left[\ddot{x}_c, \dot{x}_c, x_c\right]$ is the trajectory adjusted by admittance control in response to external force and torque.

\subsection{Implementation Details}

We employ Diffusion Policy (DP) \cite{dp} due to its ability to express multimodality and training stability. We use DDIM \cite{ddim} as the noise scheduler, with 100 steps at training and 16 steps at inference. The policy takes as input a third-view 320x240 RGB image, the end-effector 6DoF pose, and the binary gripper state. The action consists of the end-effector 6DoF pose and the binary gripper state.

\begin{table}[thpb]\tiny
\caption{Comparison results on pick-and-lift in the real world. The best results are highlighted in {\color[HTML]{FE0000} red}.}
\vspace{-0.6cm}
\label{tab: comparison_pick}
\begin{center}
\begingroup
\setlength{\tabcolsep}{6pt} 
\renewcommand{\arraystretch}{1.5} 
\resizebox{0.5\textwidth}{!}{
\begin{tabular}{lcccccccc}
\hline
            & \begin{tabular}[c]{@{}c@{}}SR-ND \\ (\%)$\uparrow$\end{tabular}                  
            & \begin{tabular}[c]{@{}c@{}}SR-D \\ (\%)$\uparrow$\end{tabular}                  
            & \begin{tabular}[c]{@{}c@{}}SR \\ (\%)$\uparrow$\end{tabular}                   
            & \begin{tabular}[c]{@{}c@{}}AT \\ (s)$\downarrow$\end{tabular}                       
            & \begin{tabular}[c]{@{}c@{}}SR-R \\ (\%)$\uparrow$\end{tabular}                   
            & \begin{tabular}[c]{@{}c@{}}GSR-ND \\ (\%)$\uparrow$\end{tabular}                 
            & \begin{tabular}[c]{@{}c@{}}GSR-D \\ (\%)$\uparrow$\end{tabular}                 
            & \begin{tabular}[c]{@{}c@{}}GSR \\ (\%)$\uparrow$\end{tabular}                   \\ \hline
DP-Real     & 20                         & 0                         & 10                        & 18.5                        & 10                         & 20                         & 0                         & 10                        \\
RDT-1B-Real & 30                         & 10                        & 20                        & 27.0                        & 70                         & 40                         & 10                        & 25                        \\ \hline
Ours        & {\color[HTML]{FE0000} 100} & {\color[HTML]{FE0000} 80} & {\color[HTML]{FE0000} 90} & {\color[HTML]{FE0000} 16.3} & {\color[HTML]{FE0000} 100} & {\color[HTML]{FE0000} 100} & {\color[HTML]{FE0000} 80} & {\color[HTML]{FE0000} 90} \\ \hline
\end{tabular}
}
\endgroup
\end{center}
\vspace{-0.3cm}
\end{table}

\begin{table}[thpb]
\caption{Comparison results on drawer-opening in the real world. The best results are highlighted in {\color[HTML]{FE0000} red}.}
\vspace{-0.6cm}
\label{tab: comparison_drawer}
\begin{center}
\begingroup
\setlength{\tabcolsep}{10pt} 
\renewcommand{\arraystretch}{1.5} 
\resizebox{0.5\textwidth}{!}{
\begin{tabular}{lcccccccc}
\hline
            & \begin{tabular}[c]{@{}c@{}}SR-ND \\ (\%)$\uparrow$\end{tabular}                  
            & \begin{tabular}[c]{@{}c@{}}SR-D \\ (\%)$\uparrow$\end{tabular}                  
            & \begin{tabular}[c]{@{}c@{}}SR \\ (\%)$\uparrow$\end{tabular}                   
            & \begin{tabular}[c]{@{}c@{}}AT \\ (s)$\downarrow$\end{tabular}                       
            & \begin{tabular}[c]{@{}c@{}}SR-R \\ (\%)$\uparrow$\end{tabular}                   
            & \begin{tabular}[c]{@{}c@{}}GSR-ND \\ (\%)$\uparrow$\end{tabular}                 
            & \begin{tabular}[c]{@{}c@{}}GSR-D \\ (\%)$\uparrow$\end{tabular}                 
            & \begin{tabular}[c]{@{}c@{}}GSR \\ (\%)$\uparrow$\end{tabular}                   \\ \hline
ACT-Real    & 0         & 0        & 0      & -     & 30       & 0          & 0         & 0       \\
DP-Real     & 50        & 0        & 25     & 24.6  & 20       & 50         & 0         & 25      \\
RDT-1B-Real & 50        & 30       & 40     & 28.0  & 30       & 50         & 30        & 40      \\ \hline
Ours        & {\color[HTML]{FE0000} 100} & {\color[HTML]{FE0000} 80} & {\color[HTML]{FE0000} 90} & {\color[HTML]{FE0000} 18.0} & {\color[HTML]{FE0000} 100} & {\color[HTML]{FE0000} 100} & {\color[HTML]{FE0000} 80} & {\color[HTML]{FE0000} 90} \\ \hline
\end{tabular}
}
\endgroup
\end{center}
\vspace{-0.3cm}
\end{table}

\begin{table}[t!]
\caption{Comparison results on oven-opening in the real world. The best results are highlighted in {\color[HTML]{FE0000} red}.}
\vspace{-0.6cm}
\label{tab: comparison_oven}
\begin{center}
\begingroup
\setlength{\tabcolsep}{10pt} 
\renewcommand{\arraystretch}{1.5} 
\resizebox{0.5\textwidth}{!}{
\begin{tabular}{lcccccccc}
\hline
            & \begin{tabular}[c]{@{}c@{}}SR-ND \\ (\%)$\uparrow$\end{tabular}                  
            & \begin{tabular}[c]{@{}c@{}}SR-D \\ (\%)$\uparrow$\end{tabular}                  
            & \begin{tabular}[c]{@{}c@{}}SR \\ (\%)$\uparrow$\end{tabular}                   
            & \begin{tabular}[c]{@{}c@{}}AT \\ (s)$\downarrow$\end{tabular}                       
            & \begin{tabular}[c]{@{}c@{}}SR-R \\ (\%)$\uparrow$\end{tabular}                   
            & \begin{tabular}[c]{@{}c@{}}GSR-ND \\ (\%)$\uparrow$\end{tabular}                 
            & \begin{tabular}[c]{@{}c@{}}GSR-D \\ (\%)$\uparrow$\end{tabular}                 
            & \begin{tabular}[c]{@{}c@{}}GSR \\ (\%)$\uparrow$\end{tabular}                   \\ \hline
ACT-Real    & 50                        & 20                        & 35                        & 34.7                        & 30                         & 50                        & 30                        & 40                        \\
DP-Real     & 60                        & 20                        & 40                        & 68.5                        & 30                         & 60                        & 20                        & 40                        \\
RDT-1B-Real & 60                        & 30                        & 45                        & 44.4                        & 60                         & {\color[HTML]{FE0000} 70}                        & 30                        & 50                        \\ \hline
Ours        & {\color[HTML]{FE0000} 70} & {\color[HTML]{FE0000} 90} & {\color[HTML]{FE0000} 80} & {\color[HTML]{FE0000} 26.7} & {\color[HTML]{FE0000} 100} & {\color[HTML]{FE0000} 70} & {\color[HTML]{FE0000} 90} & {\color[HTML]{FE0000} 80} \\ \hline
\end{tabular}
}
\endgroup
\end{center}
\vspace{-0.6cm}
\end{table}

\section{Experiments}

Our experiments are designed to answer the following questions: (1) How effective is our proposed method in disambiguating the gripper state in grasp-based tasks to improve policy robustness to gripper disturbance? (2) How does the policy trained with pure simulation data coupled with pseudo-tactile as feedback perform in real-world tasks compared to policies trained with real-world teleoperation data? (3) Can disambiguating the gripper state be achieved by extending the simulation data or changing the policy architecture? (4) Can a Vision-Language-Action (VLA) model, pre-trained on massive datasets, successfully disambiguate the gripper state after fine-tuning?

\subsection{Experiment Setups}

\textbf{Tasks.} 
We conducted experiments on three grasp-based tasks, including grasping of free objects (\textit{pick-and-lift}) and articulated objects manipulation (\textit{drawer-opening} and \textit{oven-opening}).

\textbf{Data.} 
In the real world, we used the Touch teleoperation system to collect 50 demonstrations for pick-and-lift, 60 demonstrations for drawer-opening, and 54 demonstrations for oven-opening, with each task taking approximately 3 to 4 hours. In the simulation, we used the designed privileged state-based expert policy to collect 2,000 demonstrations for each task in a similar amount of time.

\begin{figure}[thpb]
\centering
\includegraphics[width=0.48\textwidth]{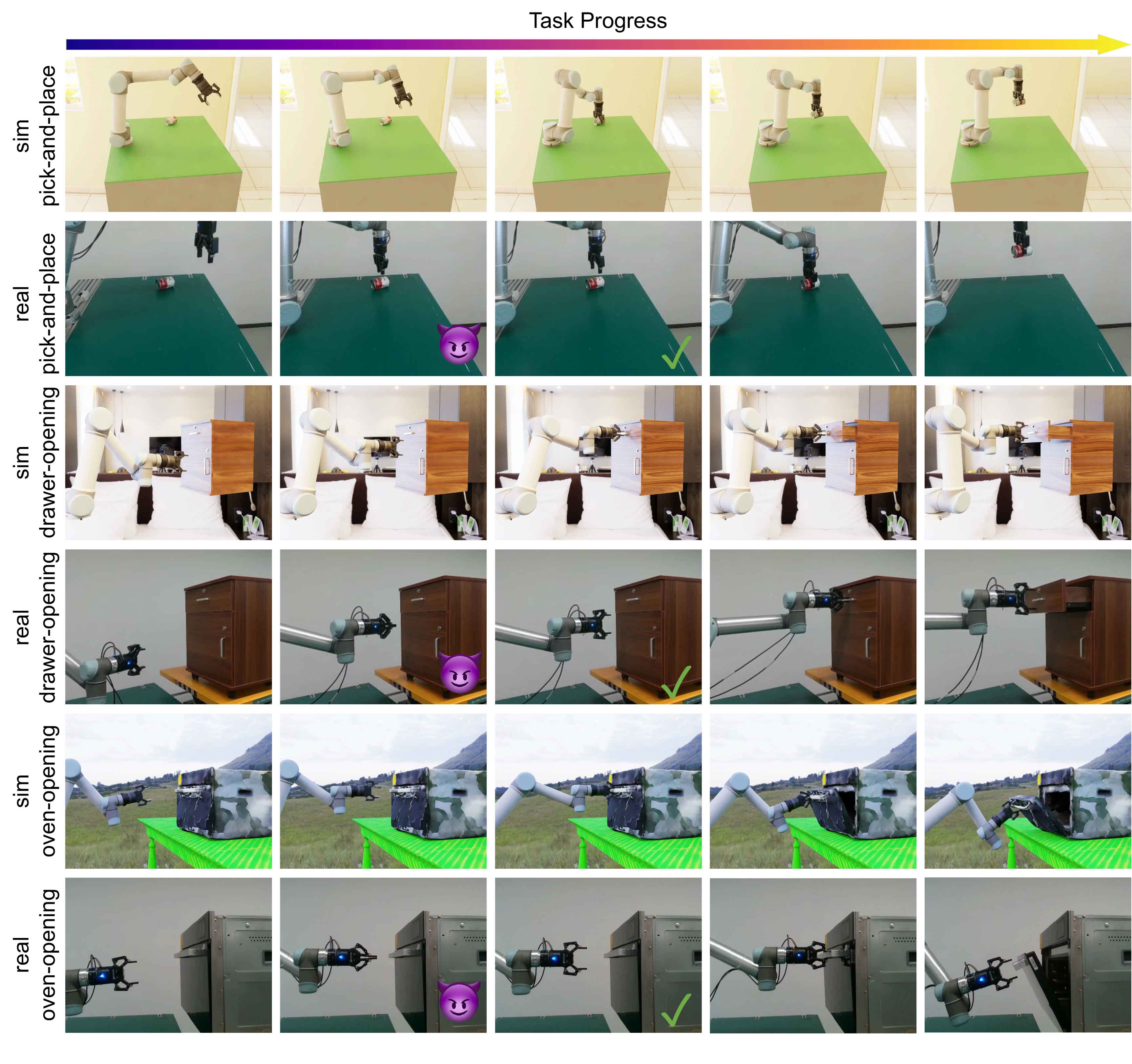}
\vspace{-0.5cm}
\caption{Visualize some rollouts of our policy both in simulation and real world.  The “little devil” symbol marks the moments when disturbances are introduced, and the checkmark symbol indicates the policy’s ability to recover from disturbance in the real world.}
\label{fig: sim&real}
\vspace{-0.5cm}
\end{figure}

\textbf{Baselines.}
We evaluate our approach against three baselines trained with real-world teleoperation data, each with different policy architectures and hyperparameters: (1) Diffusion Policy \cite{dp} with continuous gripper joint angle as observation (\textit{DP-Real}). (2) Action Chunking with Transformers \cite{act} using 640x480 RGB image (\textit{ACT-Real}). (3) Finetuned RDT-1B \cite{rdt} using 1280x720 RGB image and continuous gripper joint angle as observation (\textit{RDT-1B-Real}).

\begin{table*}[thpb]
\caption{Ablation results on oven-opening in the real world. The best results are highlighted in {\color[HTML]{FE0000} red}.}
\vspace{-0.6cm}
\label{tab: ablation}
\begin{center}
\begingroup
\setlength{\tabcolsep}{6pt} 
\renewcommand{\arraystretch}{1.5} 
\resizebox{\textwidth}{!}{
\begin{tabular}{cccccccccccc}
\hline
No. & \begin{tabular}[c]{@{}c@{}}policy \\ architecture\end{tabular} & \begin{tabular}[c]{@{}c@{}}sim data with \\ gripper randomization\end{tabular} & \begin{tabular}[c]{@{}c@{}}pseudo-tactile \\ feedback\end{tabular} & SR-ND(\%)$\uparrow$       & SR-D(\%)$\uparrow$        & SR(\%)$\uparrow$          & AT(s)$\downarrow$           & SR-R(\%)$\uparrow$         & GSR-ND(\%)$\uparrow$      & GSR-D(\%)$\uparrow$       & GSR(\%)$\uparrow$         \\ \hline
1   & DP                                                             & \textcolor{red}{\ding{55}}                                                               &\textcolor{red}{\ding{55}}                                            & 50                        & 10                        & 30                        & {\color[HTML]{FE0000} 19.7} & 30                         & 50                        & 10                        & 30                        \\
2   & DP                                                             & \textcolor{green}{\ding{51}}                                                               &\textcolor{red}{\ding{55}}                                            & 0                         & 0                         & 0                         & -                           & {\color[HTML]{FE0000} 100} & {\color[HTML]{FE0000} 70} & 80                        & 75                        \\
3   & DP                                                             & \textcolor{red}{\ding{55}}                                                               & \textcolor{green}{\ding{51}}                                            & {\color[HTML]{FE0000} 70} & {\color[HTML]{FE0000} 90} & {\color[HTML]{FE0000} 80} & 26.7                        & {\color[HTML]{FE0000} 100} & {\color[HTML]{FE0000} 70} & {\color[HTML]{FE0000} 90} & {\color[HTML]{FE0000} 80} \\
4   & ACT                                                            & \textcolor{green}{\ding{51}}                                                               & \textcolor{red}{\ding{55}}                                            & 40                        & 40                        & 40                        & 20.1                        & {\color[HTML]{FE0000} 100} & 40                        & 70                        & 55                        \\
5   & RDT-1B                                                         & \textcolor{green}{\ding{51}}                                                               & \textcolor{red}{\ding{55}}                                            & {\color[HTML]{000000} 0}  & {\color[HTML]{000000} 0}  & {\color[HTML]{000000} 0}  & {\color[HTML]{000000} -}    & {\color[HTML]{FE0000} 100} & {\color[HTML]{000000} 50} & {\color[HTML]{000000} 60} & {\color[HTML]{000000} 55} \\ \hline
\end{tabular}
}
\endgroup
\end{center}
\vspace{-0.6cm}
\end{table*}

\textbf{Metric.} 
For each task, we perform 20 rollouts in the \textit{real world}, with randomized object poses and robot initial poses for each trial. In 10 of these rollouts, the gripper is forcibly closed at different moments during the grasping phase. The metrics we use include: (1) Task Success Rate without Gripper Disturbance (\textit{SR-ND}). (2) Task Success Rate with Gripper Disturbance (\textit{SR-D}). (3) Overall Task Success Rate (\textit{SR}). (4) Average Time to Complete a Task (\textit{AT}). (5) Disturbance Resilience Success Rate (\textit{SR-R}): This metric evaluates the policy’s robustness by considering a trial successful if the policy can recover from the disturbance—by opening the gripper and continuing to attempt the grasp within a short period—regardless of whether the task is fully completed. Additionally, we tracked the performance of the grasp stage: (6) Grasp Success Rate without Gripper Disturbance (\textit{GSR-ND}). (7) Grasp Success Rate with Gripper Disturbance (\textit{GSR-D}). (8) Overall Grasp Success Rate (\textit{GSR}).

\textbf{Hardware.}
We conducted experiments on a UR5 arm with a Robotiq 2F-85 gripper. For perceptual inputs, we use an Intel RealSense D435i camera to obtain RGB images.

\vspace{-0.1cm}

\subsection{Comparisons}

As shown in Tab. \ref{tab: comparison_pick}, \ref{tab: comparison_drawer}, and \ref{tab: comparison_oven}, the policy trained with pure simulation data, coupled with pseudo-tactile feedback, outperforms the baselines trained with real-world teleoperation data across all metrics and for all three tasks in the real world. Fig. \ref{fig: qualitative_exp}(a) qualitatively illustrates the behavior for each method on the oven-opening task. We observed that baseline models often failed to cope with the disturbance (forcing the gripper to close at the moment marked by the red dashed box) due to gripper state ambiguity. In contrast, our method accomplished the task without being affected by the disturbance. Here is a detailed analysis:

\textbf{100\% Disturbance Resilience Success Rate.}
Our method achieved a perfect disturbance resilience success rate (100\%) across all tasks. This demonstrates the effectiveness of the proposed pseudo-tactile as feedback approach, which disambiguates gripper state in grasp-based tasks and enhances the policy robustness to gripper disturbance.

\textbf{Superior Task and Grasp Success Rates.}
In all three tasks, the policy trained with pure simulation data significantly outperforms the baselines trained with real-world data in both grasp success rate and task success rate, under both undisturbed and disturbed conditions. This highlights the potential of simulation data and emphasizes the significance of our proposed pseudo-tactile as feedback approach, which enables pure simulation learning by providing the policy with a noise-free binary gripper state observation.

\textbf{Higher Efficiency.}
The policy trained with pure simulation data consistently required the least amount of time to succeed across all three tasks, showcasing its efficiency. Demonstrations collected by the privileged stated-based expert policy in simulation tend to be smoother and more efficient, whereas real-world teleoperation data inevitably exhibits jitter. Additionally, the less intuitive nature of teleoperation devices leads to reduced efficiency in human demonstrations. The episode length of teleoperated demonstrations for oven-opening typically exceeds 300 steps, while most simulation demonstrations only encompass no more than 200 steps. This also indicates that the computation overhead introduced by using pseudo-tactile as feedback is minimal.

\textbf{Variation of Success Rates for Different Tasks.}
The task success rate for the baselines varied across tasks, with oven-opening performing the best, followed by drawer-opening, and pick-and-lift being the most challenging. We attribute this to the relative ease of grasping the larger oven handle, which presents fewer challenges in terms of object variability and learning complexity. In contrast, the smaller drawer handle and the highly variable nature of the pick-and-lift task increase the difficulty of learning, thereby raising the demand for larger datasets to perform effectively. In contrast, the policy trained using simulation data performs well on all three tasks because of the more abundant and diverse demonstrations. This further indicates the importance of our proposed method in making pure simulation training possible, as simulation enables large-scale data collection at a very low cost.

\textbf{Performance of Finetuned VLA Model.}
The fine-tuned RDT-1B \cite{rdt} model, a VLA model pre-trained on massive datasets, showed improved disturbance resilience compared to the from-scratch DP \cite{dp} and ACT \cite{act} baselines. However, it still underperformed when compared to our method. This suggests that scaling data and models may be effective but not always the most efficient. 

We also visualize some rollouts of our policy on the three tasks both in the simulation and the real world, shown in Fig. \ref{fig: sim&real}. Our policy can accomplish these tasks in the real world, under perturbed conditions.

\subsection{Ablations}

We conduct ablation studies on the oven-opening task to examine the effects of pseudo-tactile feedback, data, and policy architecture. Note that all policies in this section use simulation data with 320x240 RGB images and the binary gripper state. The results are shown in Tab. \ref{tab: ablation}. The third row is our implementation.

\textbf{Pseudo-Tactile Feedback is Necessary and Beneficial.}
Comparing the results of the first and third rows, the policy’s disturbance resilience success rate drops significantly without pseudo-tactile as feedback, highlighting its necessity. Moreover, incorporating pseudo-tactile as feedback improves the success rate even under undisturbed conditions. This is because the policy occasionally closes the gripper before reaching the grasp pose, but with pseudo-tactile as feedback, the policy can immediately attempt the grasp again, leading to potential success. The increased average execution time is because the metric only accounts for successful rollouts. The additional successful cases with pseudo-tactile as feedback mostly arise from disturbed scenarios or require multiple grasp attempts, which results in a longer average time.

\textbf{Data Randomization may Lead to Confusion.}
In the second row, the dataset includes cases where the gripper is closed without successfully grasping the object. The results show that while disturbance resilience is achieved, task completion is not possible. By including these empty grasp cases in the training data, the correspondence between the closed gripper state and successful grasp is broken, preventing the policy from overfitting to the untrustworthy correspondence. However, the challenge of determining grasp state is shifted to the vision. The policy needs to rely on fine-grained visual cues to assess whether the grasp was successful. The visual sim-to-real gap makes fine-grained visual cues unreliable and thus can easily lead to confusion in the real world. As shown in Fig. \ref{fig: qualitative_exp}(b), when the robot closed the gripper without grasping the object, the policy could control the gripper to open and reattempt the grasp (as shown between 1.5s and 4s), eventually succeeding in grasping the handle (as shown between 4s and 9s). However, once the robot grasps the handle, it fails to perform the expected pulling motion. As shown from 9s to 39s, the policy makes small forward and backward adjustments while repeatedly opening and closing the gripper, since it cannot reliably estimate from vision whether the robot has grasped the handle. This highlights the importance of pseudo-tactile feedback, which alleviates the challenge of grasp state estimation in simulation.

\textbf{Different Policy Architectures Help Little.}
As illustrated in the last four rows, when the dataset includes empty grasp cases, different policy architectures fail to resolve the confusion issue. These policies cannot ensure task completion while achieving disturbance resilience. Our method not only handles disturbances but also improves task completion, demonstrating the importance of pseudo-tactile as feedback in both enhancing robustness and facilitating task execution.

\addtolength{\textheight}{-0.5cm}   
                                  
\section{Conclusion}

In this work, we identified the issue of gripper state ambiguity in grasp-based tasks and provided a detailed analysis of its root cause—the absence of tactile feedback. Building on the insight that a force-controlled gripper can act as a tactile sensor, we employ pseudo-tactile as feedback to disambiguate the gripper state in grasp-based tasks. This approach enhances the policy robustness without additional data collection and hardware involvement. Moreover, it allows the policy to use the noise-free binary gripper state as an observation, enabling pure simulation learning, and thus we can benefit from the advantages of simulation. Experimental results showed that the policy trained using pure simulation data, combined with pseudo-tactile as feedback, exhibited superiority over baselines trained with real-world teleoperation data in three real-world grasp-based tasks. This underscores the effectiveness of the proposed method and highlights the potential of simulation data in training robust policies for real-world applications.


\bibliographystyle{IEEEtran}
\bibliography{IEEEabrv, references}





\end{document}